\documentclass[runningheads]{llncs}

\usepackage{eccv}

\usepackage{eccvabbrv}

\usepackage{graphicx}
\usepackage{booktabs}
\usepackage{siunitx}
\usepackage{multicol}
\usepackage{multirow}
\usepackage{wrapfig}

\usepackage[accsupp]{axessibility}  

\usepackage[pagebackref,breaklinks,colorlinks,citecolor=eccvblue]{hyperref}

\usepackage{orcidlink}

\begin{document}

\title{CoMaTrack: Competitive Multi-Agent Game-Theoretic Tracking with Vision-Language-Action Models} 

\titlerunning{Competitive Multi-Agent Game-Theoretic Tracking}

\author{Youzhi Liu \and
Li Gao\thanks{Corresponding author.} \and
Liu Liu \and
Mingyang Lv \and
Yang Cai}
\authorrunning{Y.Liu et al.}

\institute{Amap, Alibaba Group}

\maketitle

\begin{abstract}

Embodied Visual Tracking (EVT), a core dynamic task in embodied intelligence, requires an agent to precisely follow a language-specified target. Yet most existing methods rely on single-agent imitation learning, suffering from costly expert data and limited generalization due to static training environments. Inspired by competition-driven capability evolution, we propose CoMaTrack, a competitive game-theoretic multi-agent reinforcement learning framework that trains agents in a dynamic adversarial setting with competitive subtasks, yielding stronger adaptive planning and interference-resilient strategies.
We further introduce CoMaTrack-Bench, the first open-source Habitat-based benchmark protocol and episode set for language-conditioned competitive EVT featuring dynamic dueling, featuring game scenarios between a tracker and adaptive opponents across diverse environments and instructions, enabling standardized robustness evaluation under active adversarial interactions.
Experiments show that CoMaTrack achieves state-of-the-art results on both standard benchmarks and CoMaTrack-Bench. Notably, a 3B VLM trained with our framework surpasses previous single-agent imitation learning methods based on 7B models on the challenging EVT-Bench, achieving 92.1\% in STT, 74.2\% in DT, and 57.5\% in AT. The benchmark code will be available at \url{https://github.com/wlqcode/CoMaTrack-Bench}
\end{abstract}

\begin{figure}[t]
  \centering
  \includegraphics[width=\linewidth]{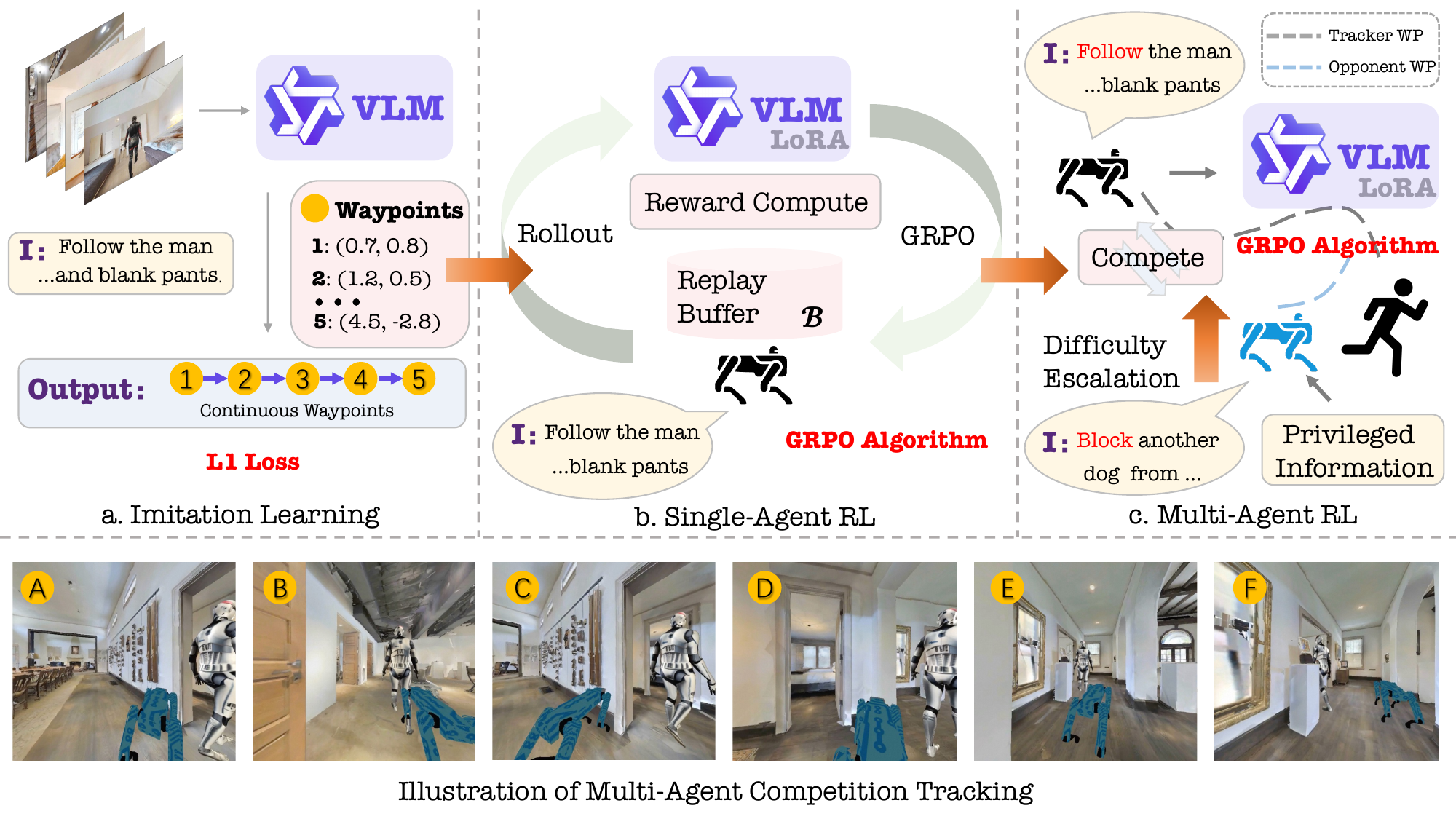}
  \caption{CoMaTrack frames EVT as a competitive multi-agent game rather than a single-agent pursuit in a static enviroment. (a) IL: the agent learns from offline demonstrations in static scenes, with limited exposure to rare failures. (b) Single-Agent RL: the agent improves through interaction, but the target and environment remain largely fixed, leading to slow exploration and overfitting to predefined behaviors. (c) Multi-Agent RL: the agent trains against adaptive opponents that evade or block on purpose, dynamically increasing difficulty and producing diverse adversarial trajectories, encouraging anticipation, relocalization, and robustness under interference.}
  \label{fig:teaser}
\end{figure}

\section{Introduction}
\label{sec:intro}
In recent years, embodied agents driven by Large Language Models (LLMs)~\cite{gpt,videollava,qwen,team2023gemini,gpt4o,kaplan2020scaling} have made remarkable progress, allowing perception, reasoning, and decision-making in complex 3D environments and moving from merely understanding instructions to executing tasks. Among them, Embodied Visual Tracking (EVT)~\cite{adcat+,follow} is a key embodied task that requires an agent to continuously follow a dynamic target from egocentric observations, while operating under environmental uncertainty, changing target behaviors, frequent occlusions, and partial observability over long horizons. Unlike static recognition or short-range navigation, EVT emphasizes persistent closed-loop control and online error correction: the agent must not only see the target, but also maintain identity consistency when the target moves, becomes occluded, is interrupted by distractors, or is confused with similar instances, and meanwhile take proper actions for pursuit and re-localization. As such, the task inherently demands target understanding, spatiotemporal reasoning, motion planning, memory maintenance, and robustness to interference—making it a critical stepping stone towards general embodied intelligence and real-world robotic applications. 

However, existing research remains largely in a single-agent setting and is predominantly trained through Imitation Learning (IL)~\cite{uninavid,racer,trackvla,dsp,liu2024navagentmultiscaleurbanstreet}, leading to limited generalization and robustness in out-of-distribution scenes. A central limitation of single-agent IL is its strong dependence on high-quality expert trajectories: data collection and annotation are expensive, and the task distribution is manually designed, making it difficult to cover the rich variations encountered in the real world~\cite{mandlekar2021matters,rt2,dasari2019robonet}.
As a result, agents trained on such a finite problem set often learn to fit the training distribution rather than acquiring transferable strategies~\cite{advat}. 

More importantly, although reinforcement learning (RL) ~\cite{chen2022reinforced,wang2024boosting,lin2021adversarial,zhong2024empowering} provides closed-loop interaction and exploration, and in principle can yield more general policies through trial-and-error~\cite{zhang2018coarse}, EVT still lacks systematic RL studies and effective training paradigms. 
Conventional RL training~\cite{ppo,deepseekmath,pi06} typically relies on relatively static environments and fixed target generators, which struggle to continuously produce hard enough and progressively more challenging supervision signals. Without an automatically escalating learning pressure, agents are prone to inefficient exploration and overfitting, resulting in limited sample efficiency and generalization gains. In other words, even with RL, if the environment itself does not become stronger, the agent is unlikely to be forced to develop robust long-horizon planning, counter-planning behaviors.

We observe that in humans and animals, capabilities often evolve rapidly through sustained competition~\cite{zhou2019survey}: changes in opponents’ strategies continuously raise task difficulty, forcing individuals to improve anticipation, planning, and counteraction~\cite{nowe2012game}. Inspired by this, we introduce CoMaTrack, a competitive multi-agent game-theoretic learning mechanism into EVT and build a co-evolving training loop among agent–opponent–environment. The agent no longer faces a static environment and a fixed data distribution; instead, it repeatedly encounters stronger, more cunning, and more diverse behaviors through adversarial interactions, yielding an automatic curriculum with progressive difficulty. Under this framework, the policies of the other agents directly reshape the training distribution—effectively making opponents the environment. Any improvement by one agent becomes a new challenge that others must overcome, ultimately forming a self-reinforcing arms race. Different from conventional multi-agent cooperation that primarily optimizes system-level efficiency, we focus on sharpening the core competence of a single tracking agent via competitive games, enabling robust behaviors under typical tracking difficulties, and improving transferability to unseen environments and target behaviors.

As shown in \cref{fig:teaser}, we shift the paradigm from static data-driven training to adversarial interaction-driven learning. Critically, existing EVT benchmarks (\eg, EVT-Bench~\cite{trackvla}, TPT-Bench~\cite{tpt}, Gym-UnrealCV~\cite{unrealcv}) exacerbate this limitation: they evaluate tracking under idealized or passively disturbed conditions, lacking scenarios with active adversarial competition that mirror real-world challenges like intentional evasion, strategic deception, or contested pursuit. To bridge this gap, we introduce CoMaTrack-Bench, the first open-source benchmark for competitive EVT, featuring dynamic dueling scenarios between a tracker agent and adaptive opponent across diverse environments and language instructions. This benchmark establishes a standardized protocol for rigorously assessing robustness under intentional adversarial interactions—moving beyond the static obstacle paradigm of prior evaluations. Our main contributions are as follows:
\begin{itemize}
    \item First integration of competitive multi-agent game-theoretic RL with modern VLA architectures for language-conditioned embodied visual tracking: We devise a co-evolving adversarial training loop where opponent strategies dynamically escalate task difficulty. Crucially, a compact 3B-parameter VLM trained under CoMaTrack surpasses all existing single-agent methods leveraging 7B-parameter models, confirming that strategic competition—not model scale—drives robust generalization.
    \item Open-sourcing CoMaTrack-Bench: We release the first open-source benchmark suite that standardizes competitive EVT evaluation in Habitat with language instructions, featuring dynamic dueling scenarios with adaptive opponents to enable rigorous, real-world-aligned evaluation of tracking robustness—addressing the idealized limitations of prior benchmarks.
    \item Extensibility to broader VLA embodied tasks: While this paper focuses on tracking, our approach offers a general paradigm of opponent-driven difficulty generation and competitive-game-driven generalization, and can be naturally extended to other VLA embodied tasks to address out-of-distribution generalization.
    
\end{itemize}

\section{Related Work}
\subsection{Visual Language Navigation}
Recent VLN~\cite{anderson2018vision,navila,etpnav,liu2024navagentmultiscaleurbanstreet} research has increasingly shifted from modular pipelines to Vision-Language-Action (VLA)~\cite{rt2,openvla,o2024open} that map egocentric observations and language instructions to actions in continuous environments. 
Uni-NaVid~\cite{uninavid} exemplifies this trend by extending video VLMs with an online token merging mechanism for efficient long-context video processing, and by scaling multi-task navigation data to enable unified navigation competence. 
DV-VLN~\cite{dvvln} introduces structured navigational chain-of-thought and a generate-then-verify procedure, using dual verification signals to re-rank candidate actions for more interpretable and dependable decisions. 
SpatialVLN~\cite{spatialvln} aims to inject explicit spatial awareness by enhancing perception through multi-sensor fusion and couples it with multi-expert reasoning and conflict-driven exploration. 
DeepVLN~\cite{deepvln} emphasizes closed-loop policy learning—adapting a LLM to VLN via supervised fine-tuning and then refining its online behavior with RL to reduce error accumulation, optionally leveraging collaborative reasoning between local and cloud models under uncertainty.

These advances collectively demonstrate that stronger VLM backbones, scalable multi-task data, explicit spatial priors, and closed-loop RL can significantly improve VLN performance. Our approach is complementary: we use multi-agent game-based RL as the core mechanism to produce robust VLA policies that can transfer to navigation-style tasks when needed, while being designed and evaluated primarily around tracking-centric requirements.

\subsection{Embodied Visual Tracking}
Embodied Visual Tracking (EVT), demanding agents to pursue language-specified targets amid dynamic occlusions, crowd interference, and environmental uncertainty, has witnessed rapid progress through large-scale imitation learning (IL) pipelines. 
TrackVLA~\cite{trackvla} pioneers a unified VLA architecture, coupling target identification and waypoint planning via supervised token prediction. 
TrackVLA++~\cite{trackvla++} enhances robustness through Polar Chain-of-Thought spatial reasoning and a confidence-gated Target Identification Memory module—yet remains fundamentally anchored to static demonstration datasets. 
Similarly, LOVON~\cite{lovon} integrates LLM-based planning, open-vocabulary perception, and a language-to-motion model for legged robot tracking; while innovating in motion-stabilized perception~\cite{aubry2014fast}, its policy derives entirely from demonstration-driven training. 
Collectively, these works epitomize the field’s paradigm: scaling EVT performance hinges on expanding expert trajectory repositories, while policies inherit critical fragilities—brittleness under distribution shift, inability to handle evasive targets, and zero capacity for strategic adaptation.


Beyond imitation-centric pipelines, reinforcement learning and game-theoretic approaches have long been explored for active visual tracking. Pioneering work such as AD-VAT~\cite{advat} introduced an asymmetric dueling mechanism where tracker and target agents engage in adversarial competition, demonstrating that co-evolutionary training yields more robust policies than single-agent paradigms. This was further theoretically analyzed in AD-VAT+~\cite{adcat+} and extended to distraction robust tracking via mixed cooperative-competitive multi-agent games~\cite{ts}. Subsequent studies validated adversarial training for tracking generalization across diverse environments~\cite{zhong2023rspt,zhong2024empowering}. However, these earlier frameworks primarily operated in non-language-conditioned settings with traditional perception backbones. HIEVT~\cite{wu2025hievt} proposes a hierarchical instruction-aware framework using LLM-based goal alignment, while ~\cite{wu2025vlm} introduces a self-improving VLM framework for embodied tracking with memory-augmented reflection mechanisms. In contrast, CoMaTrack adapts the competitive game-theoretic RL paradigm to modern VLA models, explicitly incorporating natural language instructions and leveraging large-scale VLM priors to enable robust, instruction-aware tracking under strategic adversarial interference.

\subsection{Reinforcement Learning in VLN}
Reinforcement learning (RL)~\cite{navr1,octonav,choi2017visual} has recently re-emerged as an effective tool for improving VLN agents, where supervision is often sparse, long-horizon credit assignment is difficult~\cite{richards2019dendritic}, and generalization under distribution shift remains challenging. 
OpenVLN~\cite{openvln} proposes an open-world aerial VLN framework that uses a rule-based RL procedure and a value-reward mechanism to fine-tune VLMs when training data are scarce. 
Inspired by DeepSeekR1~\cite{deepseek}, VLN-R1~\cite{vlnr1} advances RL-based alignment for continuous navigation by directly operating on egocentric video streams with large VLMs. 
Boundele \etal~\cite{bundele2024scaling} establish one of the first offline RL benchmarking suites for VLN by generating suboptimal datasets through perturbations of expert rollouts. 
ActiveVLN~\cite{activevln} performs multi-turn RL training that
treats VLN as a sequential decision process, enabling agents to learn optimal policies through direct interaction with the simulator.
While these studies demonstrate that RL can improve VLN through better reward shaping, long-horizon credit assignment, and efficient model adaptation, they remain largely confined to single-agent training regimes. In contrast, prior VLA research has not systematically adopted online RL as a primary training paradigm for EVT, although Zhong \etal~\cite{zhong2024empowering} propose an efficient EVT framework that integrates visual foundation models with offline RL and demonstrates strong sim-to-real transfer on a real robot.

We present the first GRPO-based online RL fine-tuning framework for VLA models in language-conditioned EVT, and further move beyond single-agent RL by formulating tracking as a competitive multi-agent game-theoretic training process. By leveraging competitive interactions, agents become each other’s adaptive environment, creating an automatically escalating curriculum that is difficult to obtain with static reward designs or offline data alone. 
\begin{figure}[t]
  \centering
  \includegraphics[width=\linewidth]{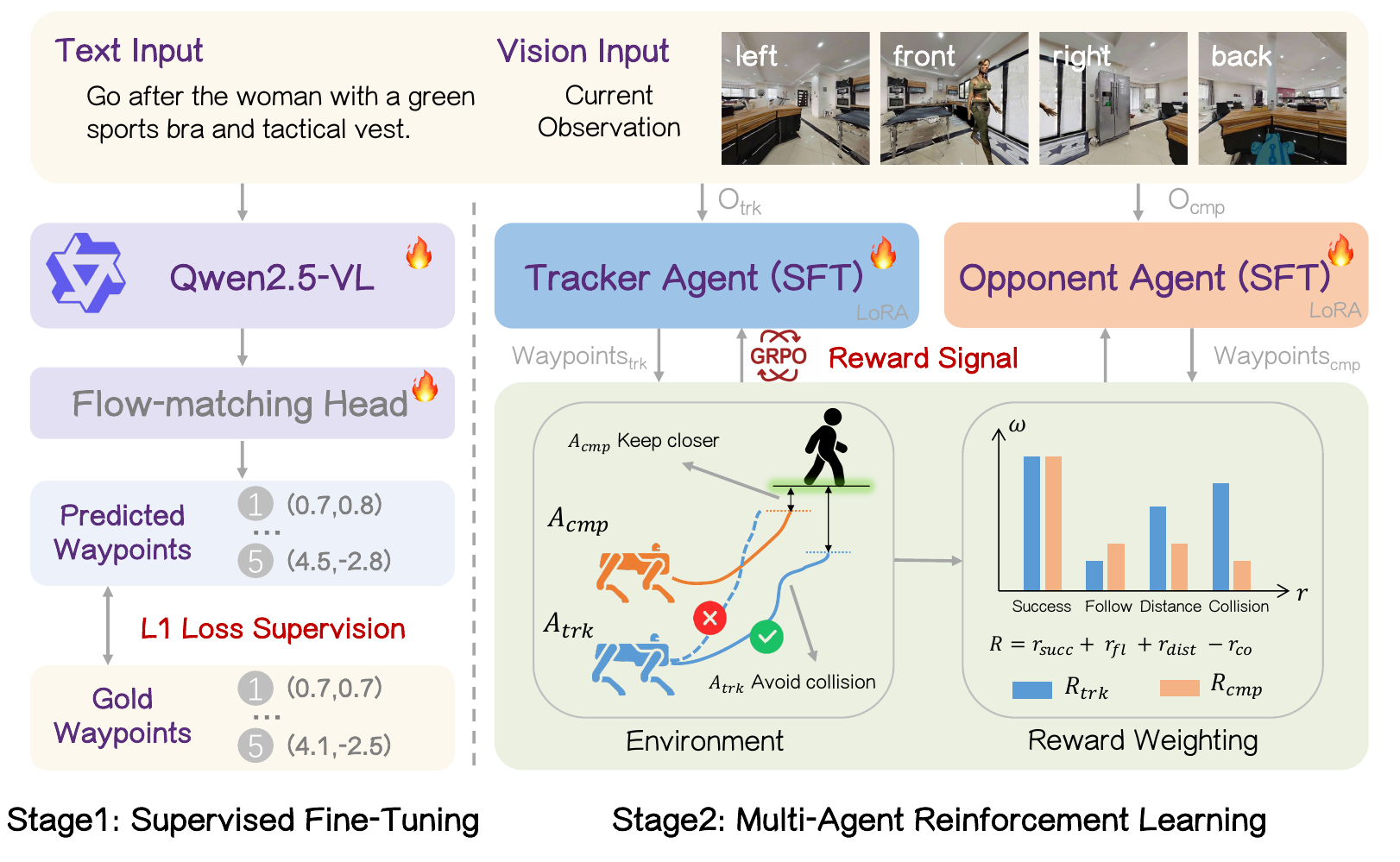}
  \caption{Overview of the CoMaTrack framework. The system employs an end-to-end VLA architecture built upon the Qwen2.5VL-3B. During the SFT phase, the model learns to predict future trajectories from multi-view observations and historical visual sequences. In the RL phase, the tracker and opponent agents engage in competitive training within a dynamic adversarial environment, co-evolving robust tracking policies through the GRPO algorithm.}
  \label{fig:method}
\end{figure}
\section{Methods}
\subsection{Task Formulation}
The EVT task can be formulated as: an agent receives a natural language instruction $\mathcal{I}$ describing the visual characteristics of the specific target together with a sequence of egocentric RGB visual observations from N cameras over the time indices $\left \{ 1,\dots,t \right \} $. Conditioned on these observations and the instruction, the agent must predict a sequence of five continuous tracking waypoints $\mathcal{W}_{T} =\left \{ w_{1},w_{2},...,w_{5} \right \} $, where each waypoint $w_{i}=(x,y,\theta) \in \mathbb{R}^3 $ specifies a relative motion in the agent’s coordinate frame, consisting of a planar displacement $(x,y)$ and a heading angle $\theta$. An episode is deemed successful when the agent follows the target at a safe distance of 1–3m, keeps the target in front of its view, and avoids collisions throughout the process.
\subsection{CoMaTrack Overview}
CoMaTrack is an end-to-end VLA model built upon a video-centric VLM backbone Qwen2.5VL-3B~\cite{qwen25}. On top of this backbone, we incorporate a flow-matching action module~\cite{flowmatching} that produces accurate, multi-modal distributions over future trajectories, enabling joint language generation and motion planning within a single framework. Text tokens are generated autoregressively in the standard manner, while trajectory planning is performed by the action module, which conditions on the backbone’s visual-linguistic representations to predict the agent’s action sequence.

\subsection{Supervised Fine-Tuning}
At time $T$, CoMaTrack processes multimodal inputs comprising current multi-view egocentric observations (front, rear, left, right) and temporally extended historical front-view sequences managed via a sliding window memory, together with a language instruction $\mathcal{I}$. Visual features are extracted using the vision backbone of Qwen2.5VL-3B~\cite{qwen25} and compressed through multi-scale grid pooling: fine-grained tokens preserve spatial fidelity of the current observation for precise target grounding, while coarse-grained tokens compactly encode long-range temporal context. The navigation-specific visual sequence is structured as $\mathbf{V}_{\text{track}} = \{ \mathbf{V}_{\text{coarse}}^{T-k}, \dots, \mathbf{V}_{\text{coarse}}^{T-4}, \mathbf{V}_{\text{fine}}^{T-3},\dots,\mathbf{V}_{\text{fine}}^{T} \}$, where $k$ denotes the memory window size. These tokens are fused with language embeddings containing a \texttt{<Nav>} special token and processed by the Qwen2.5VL-3B large language model. The LLM output adaptively branches: recognition tasks employ standard autoregressive decoding with cross-entropy loss $\mathcal{L}_{\text{text}}$; navigation tasks condition a Flow Matching-based action model~\cite{flowmatching} to directly regress a 5-step trajectory $w_i = \{(x_i, y_i, \theta_i)\}_{i=1}^{5}$, with tracking loss $\mathcal{L}_{\text{track}}$ defined as the mean squared error to ground-truth waypoints. The unified training objective is:
\[
\mathcal{L}_{\text{SFT}}
=\mathcal{L}_{\text{track}}+\alpha\,\mathcal{L}_{\text{text}}
=\sum_{i=1}^{K}\left\lVert \hat w_i-w_i\right\rVert_2^2
+\alpha\sum_{j}\mathrm{CE}\!\left(\hat t_j,t_j\right),
\]
where $\mathcal{L}_{\text{track}}$ is a waypoint regression loss,
$\mathcal{L}_{\text{text}}$ is the token-level cross-entropy loss between predicted token distributions $\hat t_j$ and ground-truth text tokens $t_j$, and $\alpha$ balances the two objectives.

\subsection{Single-Agent RL}
To overcome the limitations of IL and enable autonomous policy improvement through environmental interaction, we employ Group Relative Policy Optimization (GRPO)~\cite{deepseekmath} as our single-agent RL framework. 

\textbf{Reward Design.} To effectively guide the agent's learning, we design a  comprehensive reward function that balances dense, step-wise supervision with sparse terminal rewards. The dense rewards are engineered to provide continuous feedback on the agent's performance. This includes a distance-based reward $r_{\text{distance}}$, which follows a Gaussian distribution centered at an optimal following distance of 2.25m to encourage maintaining a safe proximity to the target:
\[
    r_{\text{distance}} = \exp\left(-\frac{1}{2}\left(\frac{d - d_{\text{opt}}}{\sigma}\right)^2\right) \cdot w_{\text{distance}}
\]
where $d$ is the current distance to the target, $d_{\text{opt}} = 2.25$, and $\sigma = 0.75$.
Additionally, we provide a facing reward for maintaining proper orientation towards the target and a tracking persistence bonus for keeping the target within a safe following zone over consecutive steps. To provide clear long-term objectives, these dense signals are complemented by sparse terminal rewards: a large positive reward for successfully completing an episode, and significant penalties for failures such as losing the target or collisions.

\textbf{GRPO Optimization.}The policy $\pi_\theta$ is optimized using the GRPO objective, which clips the probability ratio to prevent overly large policy updates:
\[
    L_{\text{GRPO}} = \mathbb{E}\left[\min\left(\frac{\pi_\theta(a|s)}{\pi_{\text{old}}(a|s)}A^{\text{group}}, \text{clip}\left(\frac{\pi_\theta(a|s)}{\pi_{\text{old}}(a|s)}, 1-\epsilon, 1+\epsilon\right)A^{\text{group}}\right)\right]
\]
where $A^{\text{group}}$ represents the group-based advantage estimate computed over trajectories of length $T_{\text{group}} = 10$. To stabilize training and prevent catastrophic forgetting of the knowledge acquired during SFT, 
we integrate a KL divergence constraint $L_{\text{KL}}$ penalizes significant deviation from the reference SFT policy $\pi_{\text{SFT}}$:
\[
    L_{\text{KL}} = \lambda_{\text{KL}} \cdot \text{KL}(\pi_\theta \| \pi_{\text{SFT}})
\]
The complete optimization objective is a weighted sum of the policy loss, an entropy bonus to encourage exploration, and the regularization terms:

\[
    L_{\text{total}} = L_{\text{GRPO}} - \lambda_{\text{ent}}\mathcal{H}(\pi_\theta) + L_{\text{KL}}
\]

\subsection{Multi-Agent RL}
To cultivate robust tracking policies under active adversarial interference, we extend the single-agent RL framework into a competitive multi-agent game-theoretic RL paradigm. Specifically, we deploy two quadrupedal agents: the tracker agent $A^{trk}$, which is required to follow the language-specified human target, and the opponent agent $A^{cmp}$, which competes for the same target and implicitly interferes with the tracker through occlusion, path crossing, and physical blocking.  Both agents adopt the identical VLA architecture and optimization pipeline. Critically, their policies are initialized from the SFT-trained checkpoint, ensuring strong behavioral priors before RL refinement. However, their reward functions are asymmetrically designed to reflect distinct strategic objectives, driving co-evolutionary policy adaptation.

\textbf{Asymmetric Reward Design.}The opponent should aggressively approach the target to create contested tracking, while the tracker should maintain stable following and avoid collisions with the opponent. Let $d_{\text{trk}}$ and $d_{\text{cmp}}$ denote the distances from $A^{\text{trk}}$ and $A^{\text{cmp}}$ to the target, and let $d_{\text{int}}$ be the inter-agent distance between the two robots.

\textbf{Opponent reward $R^{\text{cmp}}$.}We encourage the opponent to track at a closer distance than the nominal safe-following range. We use a dense Gaussian distance reward but shift the optimum to a nearer value $d^{\text{cmp}}_{\text{opt}}=1.25m$  to produce more direct contention.

\textbf{Tracker reward $R^{\text{trk}}$.}The tracker retains the standard EVT objective in Sec.3.4, but we additionally incorporate an opponent-aware safety term to discourage collisions and unsafe proximity to $A^{\text{cmp}}$.

\textbf{Multi-Agent GRPO Optimization.}Given simultaneous rollouts under the joint environment dynamics, we optimize each agent's policy with the same GRPO algorithm, but using its own trajectory returns.

\subsection{CoMaTrack Benchmark}
To systematically evaluate tracking robustness under active adversarial interactions and bridge the realism gap in existing benchmarks, we introduce CoMaTrack-Bench, the first open-source benchmark suite that standardizes competitive EVT evaluation in Habitat with language instructions. Unlike prior benchmarks that evaluate tracking under idealized or passively disturbed conditions, CoMaTrack-Bench features dynamic dueling scenarios where a tracker agent must pursue a language-specified target while contending with adaptive opponents that actively interfere with the tracking task.

\textbf{Benchmark Construction.} CoMaTrack-Bench is built upon the Single-Target Tracking (STT) data from EVT-Bench, inheriting its diverse environments from HM3D~\cite{hm3d} and MP3D~\cite{mp3d} and natural language target descriptions. To create competitive scenarios, we augment each STT episode by introducing a second robotic dog agent initialized 0.5 meters ahead of the tracker's starting position, ensuring immediate interaction from episode initialization.

\textbf{Opponent Behavior Taxonomy.} We design three progressively challenging opponent behaviors to comprehensively assess tracking robustness: (1) Static Obstacle: The opponent remains fixed at its initial position, serving as a static obstacle that occludes the target or blocks pursuit paths. This tests the tracker's ability to handle persistent spatial occlusions and path planning around stationary obstacles. (2) Random Interference: The opponent executes random patterns within the environment, creating unpredictable dynamic occlusions and path crossings. This evaluates robustness to stochastic disturbances and the tracker's capacity to maintain target identity amid moving distractors. (3) Competitive Tracking: The opponent loads the same SFT-trained policy as the tracker and actively competes to follow the same target, creating direct adversarial pursuit scenarios where both agents vie for optimal tracking positions.


\section{Training Recipe and Data Collection}

\subsection{Data Collection}
To improve the model’s versatility and generalization, and to facilitate transfer to other VLN-style tasks, we train not only on EVT data but also on navigation data collected from diverse synthetic environments (\eg, HM3D~\cite{hm3d}). In addition, we incorporate large-scale real-world VQA data~\cite{llava} to further strengthen cross-domain generalization.

\textbf{Tracking Data Collection.}
Following TrackVLA~\cite{trackvla}, we collect EVT training data using EVT-Bench~\cite{trackvla} built on Habitat 3.0~\cite{habitat3} with the scenes from HM3D~\cite{hm3d} and MP3D~\cite{mp3d}. In total, we gather 6913 Single-Target Tracking (STT) episodes, 6685 Distracted Tracking (DT) episodes, and 6524 Ambiguity Tracking (AT) episodes.

\textbf{Question Answering Data Organization.}
To equip the model with real-world 3D spatial reasoning, we incorporate ScanQA~\cite{scanqa}. We further leverage LLaVA-Pretrain~\cite{llava} to enhance general visual-language understanding. To broaden and diversify representation learning across wider domains, we additionally incorporate SYNTH-PEDES~\cite{SYNTH-PEDES}. Finally, we employ RefCOCO~\cite{refcoco} and Flickr30k~\cite{flickr30k} to jointly strengthen grounded language understanding and image captioning capabilities.
\textbf{Multi-Task Navigation Data Collection.}
We collect multi-task navigation data from diverse sources. First, we incorporate instruction-following data from YouTube videos like NaViLA~\cite{navila} and use a shortest-path follower in the Habitat simulator~\cite{habitat3} to generate expert action sequences for R2R-CE~\cite{r2r} and RxR-CE~\cite{rxr}.
Second, we collect object-goal navigation data from HM3D ObjectNav~\cite{hm3d} and HM3D OVON~\cite{ovon}.
\subsection{Training Recipe}
Our training framework adopts a principled two-phase optimization strategy:
First, following established VLM alignment protocols~\cite{llava}, we perform supervised pre-training for only one epoch optimization of the projector, vision encoder, and LLM parameters using the curated navigation, tracking and VQA datasets. 
Second, we then fine-tune the model for one epoch using GRPO~\cite{deepseekmath} under our multi-agent competitive training setup. In this stage, we keep the backbone largely fixed and apply LoRA~\cite{lora} adapters to the LLM, which are the only trainable parameters during RL fine-tuning.
\section{Experiments}
\subsection{Experiment Setups}
\textbf{Benchmarks.}

To substantiate the robustness and generalization capability of our framework, we perform rigorous validation across two evaluation protocols: three tasks in EVT-Bench~\cite{trackvla}, and CoMaTrack-Bench, a novel benchmark introduced in this work. In parallel, an extensive ablation study positions our method against current state-of-the-art approaches, such as IBVS~\cite{ibvs}, DiMP~\cite{dimp}, SARL~\cite{sarl}, AD-VAT~\cite{advat}, AD-VAT+~\cite{adcat+}, TS\cite{ts}, EVT~\cite{zhong2024empowering}, PoliFormer~\cite{poliformer}, Uni-NaVid~\cite{uninavid}, VLingNav~\cite{vlingnav}, TrackVLA~\cite{trackvla}, NavFoM~\cite{navform} and TrackVLA++~\cite{trackvla++}. This multi-faceted comparison ensures a holistic assessment of performance across diverse algorithmic paradigms and task formulations.

\textbf{Metrics.}
Tracking performance is quantitatively assessed using the canonical metric suite endorsed by the EVT-Bench~\cite{trackvla}. Specifically, we report the success rate (SR), tracking rate (TR), and collision rate (CR).

\begin{wrapfigure}{r}{0.48\textwidth}
  \centering
  \vspace{-20pt}
  \includegraphics[width=0.46\textwidth]{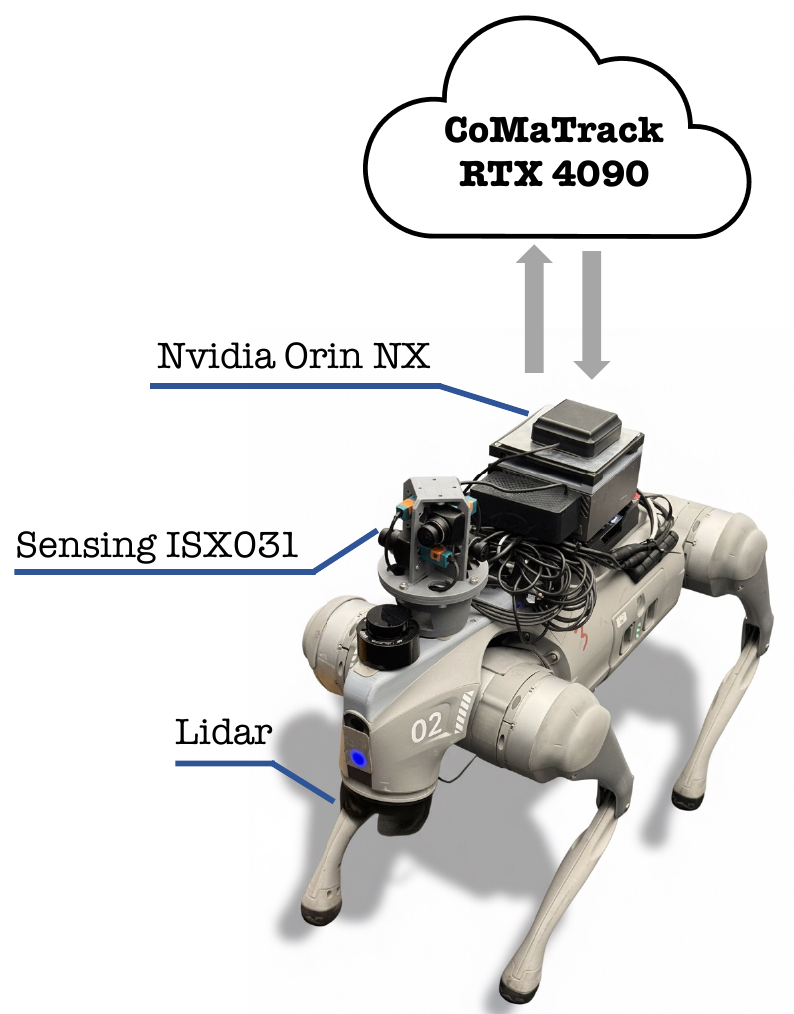}
  \caption{Hardware Platform. Our deployment platform is built on a Unitree Go2 X quadrupedal robot equipped with four monocular RGB cameras, a Unitree 4D LiDAR L2.}
  \vspace{-50pt}
  \label{fig:dep}
\end{wrapfigure}

\textbf{Implementation Details.} 
During supervised fine-tuning, CoMaTrack is trained for one epoch on a cluster of 48 NVIDIA H20 GPUs using the full set of navigation and VQA data. Following Uni-NaVid~\cite{uninavid}, we prepend a task indicator token \texttt{⟨NAV⟩} for both the EVT and navigation tasks. In the subsequent multi-agent RL stage, we train CoMaTrack on 4 NVIDIA L20 GPUs for one epoch, using only tracking data.

\textbf{Real World Deployment.}
CoMaTrack operates on a Unitree GO2 X robot equipped with four Sending ISX031 cameras during real-world deployment. The video stream is transmitted to a remote server powered by an NVIDIA RTX 4090 GPU for processing. After model inference is completed, the results are transmitted back to the local device via the network, as shown in \cref{fig:dep}

\subsection{Benchmark Results}
\begin{table}[t]
    \centering
    \caption{\textbf{Performance on EVT-Bench. Bold} denote the best results. 
    $\ast$ means using GroundingDINO~\cite{groundingdino} as the detector. 
    $\ddagger$ means using SoM~\cite{som} and GPT-4o~\cite{gpt4o} as the visual foundation model. 
    Models annotated with $\dagger$ were fine-tuned on a 7B VLM, whereas our method is built upon a \textbf{3B VLM.}}
    \label{tab:evt-bench}
    \small
    \setlength{\tabcolsep}{4pt} 
    \begin{tabular}{l*{9}{c}}
        \toprule
        \multirow{2}{*}{\textbf{Methods}} 
            & \multicolumn{3}{c}{\textbf{STT}} 
            & \multicolumn{3}{c}{\textbf{DT}} 
            & \multicolumn{3}{c}{\textbf{AT}} \\
        \cmidrule(lr){2-4} \cmidrule(lr){5-7} \cmidrule(lr){8-10}
         & SR$\uparrow$ & TR$\uparrow$ & CR$\downarrow$ 
         & SR$\uparrow$ & TR$\uparrow$ & CR$\downarrow$ 
         & SR$\uparrow$ & TR$\uparrow$ & CR$\downarrow$ \\
        \midrule
        IBVS$\ast$~\cite{ibvs}  
            & 42.9 & 56.2 & 3.8 & 10.6 & 28.4 & 6.1 & 15.2 & 39.5 & \textbf{4.9} \\
        PoliFormer$\ast$~\cite{poliformer}   
            & 4.7 & 15.5 & 40.1 & 2.6 & 13.2 & 44.5 & 3.0 & 15.4 & 41.5 \\
        EVT~\cite{zhong2024empowering} 
            & 24.4 & 39.1 & 42.5 & 3.2 & 11.2 & 47.9 & 17.4 & 21.1 & 45.6 \\
        EVT$\ddagger$~\cite{zhong2024empowering}     
            & 32.5 & 49.9 & 40.5 & 15.7 & 35.7 & 53.3 & 18.3 & 21.0 & 44.9 \\
        Uni-NaVid$\dagger$~\cite{uninavid}     
            & 53.3 & 67.2 & 12.6 & 31.9 & 50.1 & 21.3 & 15.8 & 41.5 & 26.5 \\
        TrackVLA$\dagger$~\cite{trackvla}     
            & 85.1 & 78.6 & 1.7 & 57.6 & 63.2 & 5.8 & 50.2 & 63.7 & 17.1 \\
        VLingNav$\dagger$~\cite{vlingnav}     
            & 88.4 & 81.2 & 2.1 & 67.7 & 73.5 & 5.5 & -- & -- & -- \\
        NavFoM$\dagger$~\cite{navform}     
            & 88.4 & 80.7 & -- & 62.0 & 67.9 & -- & -- & -- & -- \\
        TrackVLA++$\dagger$~\cite{trackvla++}     
            & 90.9 & 82.7 & 1.5 & 74.0 & 73.7 & 3.5 & 55.9 & 63.8 & 15.1 \\
        \textbf{Ours}   
            & \textbf{92.1} & \textbf{90.3} & \textbf{0.9} 
            & \textbf{74.2} & \textbf{80.5} & \textbf{2.1} 
            & \textbf{57.5} & \textbf{73.4} & 12.0 \\
        \bottomrule
    \end{tabular}
\end{table}

\textbf{Performance on EVT-Bench.}
We first evaluate our method on the public benchmark EVT-Bench~\cite{trackvla}, with results reported in \cref{tab:evt-bench}. 
Compared with the strongest prior baseline TrackVLA++, CoMaTrack improves SR from 90.9 to 92.1 on STT, from 74.0 to 74.2 on DT, and from 55.9 to 57.5 on AT. In addition, CoMaTrack achieves consistently higher TR and lower CR across the three tasks, indicating improved tracking persistence and safety under distractors and ambiguity.
Notably, our 3B model achieves state-of-the-art results against all existing 7B baselines, underscoring the effectiveness of the proposed multi-agent competition RL training paradigm.

\textbf{Performance on CoMaTrack-Bench.}
To rigorously evaluate tracking robustness under active adversarial interactions, we assess CoMaTrack on our newly introduced CoMaTrack-Bench. As shown in Tab.~\ref{tab:comatrack-bench}, our method substantially outperforms the baseline approach across all metrics. We compare against Uni-Navid~\cite{uninavid} as it is the only baseline method with publicly available model weights, enabling direct evaluation on our benchmark. 

Notably, CoMaTrack-Bench presents significantly more challenging scenarios compared to EVT-Bench, as it features dynamic adversarial opponents that actively interfere with tracking through occlusion, path blocking, and competitive pursuit—unlike the static or passively disturbed environments in EVT-Bench. Consequently, the absolute performance metrics on CoMaTrack-Bench are lower than those on EVT-Bench, reflecting the increased difficulty inherent in competitive tracking scenarios where the tracker must contend with strategic, adaptive interference.

\begin{table}[t]
\centering
\begin{minipage}[t]{0.48\columnwidth}
    \centering
    \caption{Zero-shot Performance on CoMaTrack-Bench.}
    \label{tab:comatrack-bench}
    \setlength{\tabcolsep}{6pt} 
    \begin{tabular}{lccc}
        \toprule
        \textbf{Methods} & SR$\uparrow$ & TR$\uparrow$ & CR$\downarrow$ \\
        \midrule
        Uni-Navid & 42.4 & 56.5 & 23.8 \\
        \textbf{Ours} & \textbf{85.0} & \textbf{82.9} & \textbf{5.5} \\
        \bottomrule
    \end{tabular}
\end{minipage}\hfill
\begin{minipage}[t]{0.48\columnwidth}
    \centering
    \caption{Ablation Study of Multi-Agent.}
   \label{tab:ablation} 
   \setlength{\tabcolsep}{6pt} 
    \begin{tabular}{l*{3}{c}}
        \toprule
         \textbf{Methods} & SR$\uparrow$ & TR$\uparrow$ & CR$\downarrow$ \\
        \midrule
        SFT Model   & 88.2 & 85.4  & 3.1 \\
        Single-Agent RL   & 89.5 & 88.0  & 2.2 \\
        Multi-Agent RL   & \textbf{92.1} & \textbf{90.3}  & \textbf{0.9} \\

        \bottomrule
    \end{tabular}
\end{minipage}
\end{table}

\subsection{Ablation Study of Multi Agents}
\begin{wrapfigure}{r}{0.48\textwidth}
  \centering
  \vspace{-6pt}
  \includegraphics[width=0.46\textwidth]{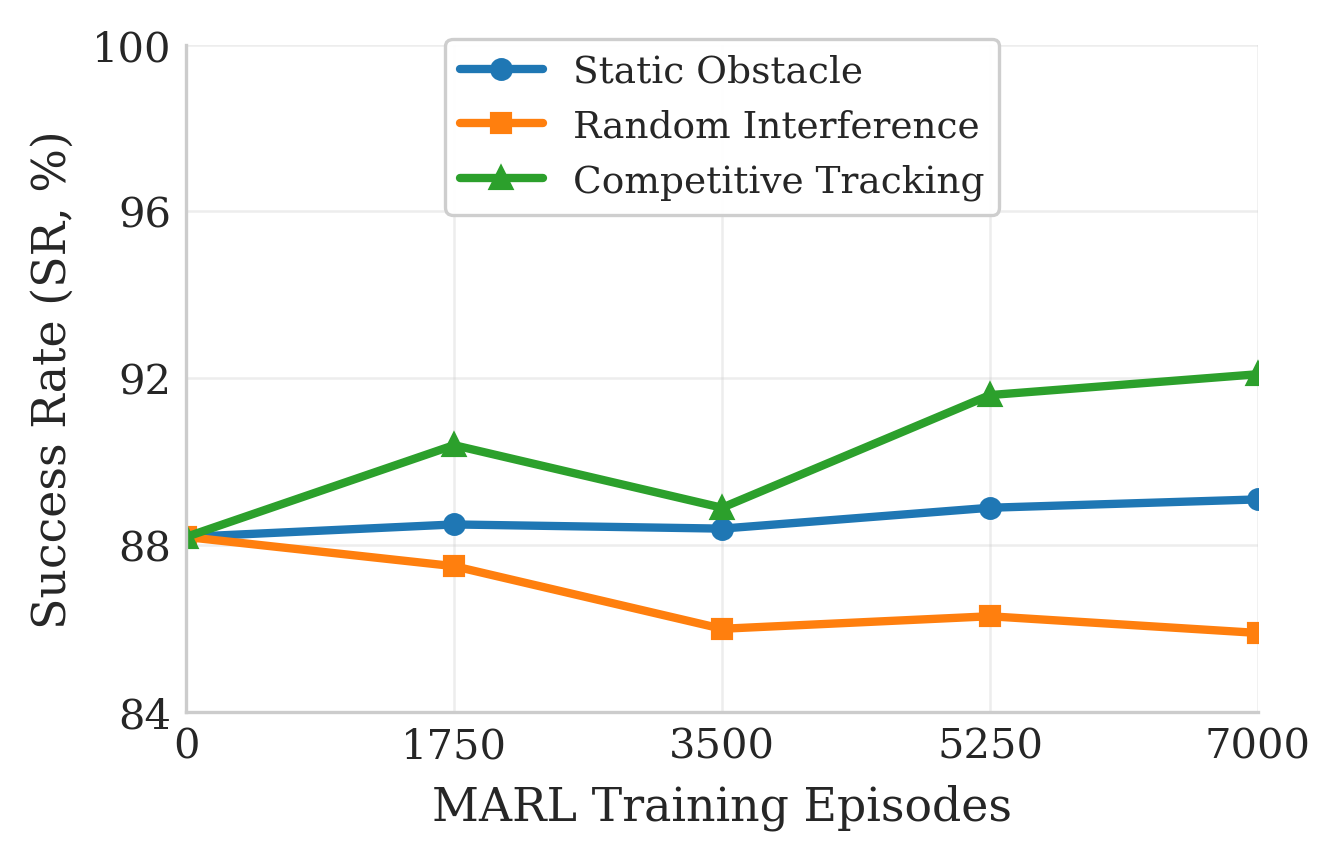}
  \caption{Diagram Illustrating the Impact of Opponent Strength on Outcomes.}
  \vspace{-10pt}
  \label{fig:marl}
\end{wrapfigure}
To quantify the contribution of our proposed multi-agent RL competitive training paradigm, we conduct a comprehensive ablation study comparing three training configurations: (1) the SFT-only baseline, (2) single-agent RL fine-tuning, and (3) our full multi-agent RL framework. As presented in Tab.~\ref{tab:ablation}, each component progressively enhances performance across all metrics.

The SFT Model trained solely on expert demonstrations achieves 88.2\% SR, 85.4\% TR, and 3.1\% CR, establishing a strong behavioral prior but lacking mechanisms for handling distribution shift or adversarial interference. Introducing Single-Agent RL improves SR to 89.5\%, TR to 88.0\%, and reduces CR to 2.2\%, demonstrating that closed-loop policy optimization through environmental interaction yields moderate gains in robustness and safety. However, the improvements remain limited.

In contrast, Multi-Agent RL achieves 92.1\% SR, 90.3\% TR, and notably reduces CR to just 0.9\%. 
These results confirm that competitive multi-agent training, where adaptive opponents continuously escalate task difficulty through strategic interference, creates a self-reinforcing curriculum that drives the emergence of robust, anticipatory tracking behaviors. The dramatic collision reduction particularly highlights that the agent learns sophisticated spatial reasoning and safe maneuvering strategies when forced to navigate around intelligent, non-cooperative opponents—capabilities difficult to acquire through static demonstration data or single-agent exploration alone.

To better understand how competitive training improves robustness, we further analyze the success rate as a function of the number of multi-agent RL training episodes under the three opponent settings in CoMaTrack-Bench. As shown in Fig.~\ref{fig:marl},Static Obstacle brings only marginal gains, while Random Interference even leads to performance degradation, indicating that stochastic disturbances alone do not provide a sufficiently structured curriculum for learning interference-resilient behaviors. The competitive tracking strategy achieves the best performance, demonstrating that stronger opponents create more informative training signals and more effectively drive policy improvement.

\begin{figure}[t]
  \centering
  \includegraphics[width=\linewidth]{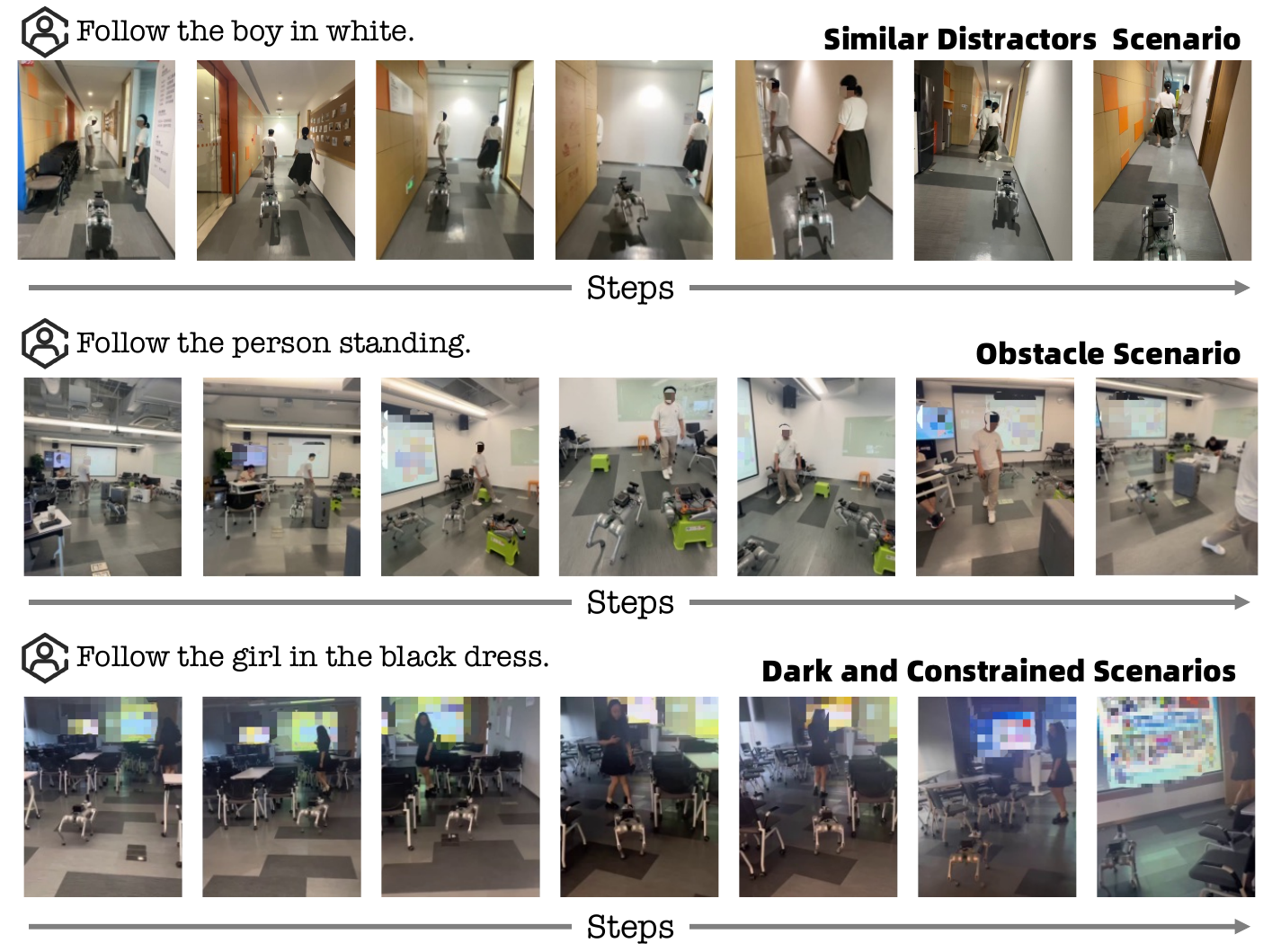}
  \caption{Qualitative real-world results demonstrating CoMaTrack’s zero-shot deployment capabilities.}
  \label{fig:realworld}
\end{figure}
\subsection{Qualitative Results in Real-World}
~\cref{fig:realworld} presents qualitative real-world evaluations of our method under challenging conditions, including (A) similar distractors scenario, (B) obstacle scenario, and (C) dark and constrained environments. The results show that CoMaTrack transfers effectively from simulation to the real world for EVT, supporting zero-shot deployment in highly dynamic settings.
\section{Conclusion}
This paper addresses two core challenges in EVT: the weak generalization of single-agent IL and the lack of effective RL training paradigms. We propose CoMaTrack, the first EVT training framework that integrates multi-agent competitive game-theoretic with RL. By constructing a co-evolution loop between a tracker and adaptive opponents, CoMaTrack shifts EVT training from a static, data-driven regime to an opponent-driven adversarial learning process, substantially improving tracking success. We further release CoMaTrack-Bench, the first competitive EVT benchmark, pushing evaluation beyond idealized static settings toward realistic adversarial scenarios.

Our experiments demonstrate both effectiveness and practicality. Multi-agent strength-controlled studies show a clear step-wise performance improvement as the opponent is upgraded from a static obstacle to random interference and further to a competitive tracking agent. Under the same evaluation protocol, a 3B-parameter VLM trained with CoMaTrack significantly outperforms all prior single-agent methods relying on 7B-parameter models, indicating that competition-driven policy evolution, rather than model scale, is the key driver of robust generalization. 

Overall, CoMaTrack provides an efficient and transferable solution for EVT and establishes a new embodied learning paradigm of game-theoretic generalization, whose central idea naturally extends to broader VLA embodied tasks such as instruction following and object-goal navigation.
\section{Limitations}
While our multi-agent game-theoretic RL framework shows strong performance on EVT, several limitations remain. The current validation focuses on EVT and its competitive setting, without large-scale evaluation on broader VLN tasks such as instruction following, object navigation, limiting demonstrated task generality. Opponent strategies, though diverse, are bounded by simulated priors and may not reflect real-world dynamics, risking distribution shift. Training is computationally costly and unstable due to multi-agent non-stationarity, requiring further optimizations in sampling and stabilization. Future work will extend the framework to more tasks, enhance opponent generation, improve training efficiency, and refine evaluation to better reflect long-term performance.

\bibliographystyle{splncs04}
\bibliography{main}
\end{document}